# Bayesian Learning of Consumer Preferences for Residential Demand Response

**Mikhail V. Goubko*, \*\*. Sergey O. Kuznetsov. Alexey A. Neznanov. Dmitry I. Ignatov. \*\*\***

*V.A. Trapeznikov Institute of Control Sciences of Russian Academy of Sciences, Moscow, Russia (e-mail: mgoubko@mail.ru).
\*\* Skoltech Center for Energy Systems, Skolkovo Institute of Science and Technology, Moscow, Russia.
\*\*\* National Research University Higher School of Economics, Moscow, Russia

**Abstract:** In coming years residential consumers will face real-time electricity tariffs with energy prices varying day to day, and effective energy saving will require automation – a recommender system, which learns consumer's preferences from her actions. A consumer chooses a scenario of home appliance use to balance her comfort level and the energy bill. We propose a Bayesian learning algorithm to estimate the comfort level function from the history of appliance use. In numeric experiments with datasets generated from a simulation model of a consumer interacting with small home appliances the algorithm outperforms popular regression analysis tools. Our approach can be extended to control an air heating and conditioning system, which is responsible for up to half of a household's energy bill.

## 1. INTRODUCTION

According to Farhangi (2010), liberalization of electricity markets is an important trend in the development of electric power systems worldwide. One of the goals is to increase elasticity of demand by reallocating risks to end customers (Albadi and El-Saadany (2008), Chan et al. (2012)). Soon not only commercial but also residential consumers will face dynamic (and, perhaps, fluent) pricing schedules reflecting supply and demand balance at the regional market, see Ipakchi and Albuyeh (2009).

Under real-time pricing the electricity saving process becomes tricky. With fixed time-of-use tariffs a rational scenario for repeating actions (conditioning, cooking, washing, etc.) can be chosen once for all. Otherwise, when prices change every day, an optimal scenario varies from day to day taking high efforts to make a rational choice. Mohsenian-Rad and Leon-Garcia (2010) argue that the elasticity of residential demand can be increased only under sufficient level of appliance control automation. Such control is a part of integral home automation following the concept of "*smart home*".

The standard utility-based approach assumes that comfort has its price: when the electricity price grows, a consumer wishes to limit her comfort level and save money. Rational appliance control reduces to balancing energy bill and consumer's comfort level, and revelation of implicit comfort level from consumer's actions becomes the key problem solved with advanced machine learning techniques (Downey (2013), Murphy (2012)). It is understood that real-life human actions are often impulsive and, to some extent, irrational, so statistical learning methods (namely, the Bayesian framework) becomes a relevant tool to learn personal preferences over scenarios of home appliance use. Then learned preferences are used in algorithms of automatic home appliance control.

The contribution of the present paper is a general methodology, which allows studying usage scenarios of a wide range of home appliances on the basis of computer simulations of consumer's rational behavior. Simulations aim to provide a realistic and compact parameterization of consumer's comfort level function, which is then used to design efficient preference learning algorithms. The same simulations are used to generate source data for algorithm testing. We illustrate our approach on a class of small-scale home appliances and suggest a Bayesian learning algorithm, which outperforms contemporary regression analysis tools in accuracy of consumer's action prediction.

## 2. PROBLEM SETTING

### 2.1 Breadmaker Case Study: Consumer's View

The following case study is widely used below to illustrate the general ideas. A *breadmaker* is used to bake bread at home. Basically, an automated baking program includes the sort of bread to make and the desired finish time. Baking is time-consuming, so to have fresh bread on time a consumer has to plan appliance usage in advance.

Typically, breadmaker usage is triggered by bread stock scarcity. Most breadmaker users prefer freshly baked bread and coordinate the finish of the program with the chow time. On the other hand, night electricity prices are lower and baking at night is the cheapest scenario. The program finish time is chosen to balance the delight of having the hot bread for the breakfast with savings from the night tariff (EUR 50-65 yearly economy is reported by Gottwalt et al. (2011) for a similar appliance). When prices vary from day to day, the trade-off becomes non-trivial, requiring support and automation.

When choosing a program, a consumer should be warned about the cost of each alternative (the sort of bread and the program finish time) to be able to weigh up all options. Under the further level of automation the system recommends the best finish time basing on the predicted electricity prices and the current awareness of consumer's preferences. The consumer agrees or chooses another scenario. A possible mockup of breadmaker UI is shown in Fig. 1. Below we consider in detail the program finish time recommendation problem.

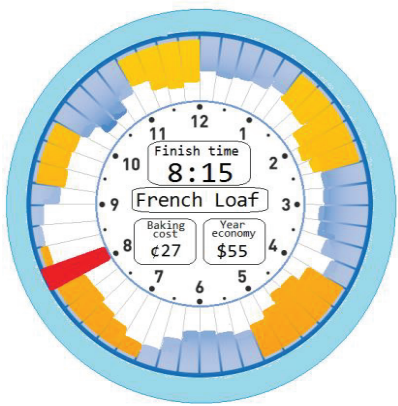

Fig. 1. A mockup of UI for automated breadmaker control. Display is surrounded by the "roll-and-press" control wheel. Baking program, finish time, baking cost, and yearly economy are shown. Alternative scenario costs are plotted around with current finish time marked red and recommended time slots marked yellow.

### 2.2 Behind the Scene

To help a consumer with his or her breadmaker, several problems must be solved by the smart home environment. The main components of the system are the preference learning module and the recommender system.

The learning module collects the statistics of consumer's actions together with the history of the system state, reveals and corrects the model of consumer's preferences. The current estimate of preferences is passed to the recommender / automatic control module, where it is combined with the current environment state to give the recommended scenario of appliance use or an automatic control action. The consumer follows the recommendation or selects a custom action, and the learning data is updated.

Measuring the consumer's comfort level function makes the key problem. Possibility of direct comfort evaluation (through questionnaires, etc.) is questionable, so preferences should be learned from consumer's actions.

### 2.3 Formal Preference Learning Problem

Under the utility-based approach the problem of optimal appliance control reduces to trading-off energy bill and comfort level by choosing a scenario of appliance use:

$$s^*(z) = \mathrm{argmax}_{s \in A} u(s, z), \quad (1)$$

where $A$ is the set of available scenarios, $z = (y, \omega)$ is the vector of relevant attributes of the current situation with vector $y$ of publicly observed components and the vector $\omega$ of hidden or *private* components observed by the consumer. Then, $u(s, z) = d(s, z) - c(s, y)$ is the utility function, where $c(s, y)$ is the total electricity cost for scenario $u$, $d(s, z)$ is the comfort level in situation $z$ under scenario $s$.

In case of a breadmaker, $A$ is a set of possible combinations of the sort of bread and the finish time. The vector $y$ might include current time and day of the week, time passed from the previous breadmaker run, and so on. Knowing the price forecast and a power consumption profile of all baking programs (see Figure 2 for a typical example) it is an easy job to write the cost $c(s, y)$ for any scenario $s \in A$ (i.e., for any combination of the sort of bread and the finish time).

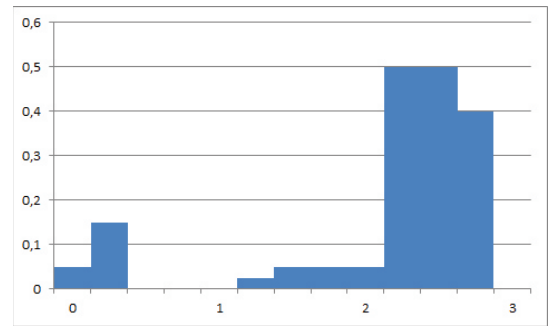

Fig. 2. Power consumption (kW) vs time (hs) for a fast baking program from Lopez (2005).

The vector $\omega$ contains future bread consumption (time and volume schedule), which may be realized by the consumer (but not the external observer) and affect the valuation of the future level of satisfaction $d(s, z)$ from bread consumption. Only general information is available on $d(s, z)$ (e.g., it will possibly increase if a freshly baked loaf is available for the next meal) and its determination is the main goal of the preference learning process.

The data available is the history of breadmaker use with each observation $i = 1, \ldots, n$, being a single breadmaker run, i.e., the tuple $(y_i, c_i(s), s_i^*)$, where $y_i$ is the public part of the system situation vector, $c_i(s)$ is the cost function for all scenarios $s \in A$, and $s_i^*$ is a real consumer choice. The problem is to predict the future action $s^*(y, \cdot)$ for any given observed situation $y$ and costs $c(s, y)$.

A direct approach assumes learning the action from the data with regression analysis, but to take advantage of rationality of consumer's choice (expression (1)) one has to learn instead an unknown function $d(s, z)$ from the history $(y_i, c_i(s), s_i^*)$ of choices (which are assumed rational).

# 3. LITERATURE REVIEW

## 3.1 Automatic Control of Home Appliances

As noted in Mohsenian-Rad and Leon-Garcia (2010), residential electricity saving under real-time electricity prices demands high level of home automation. Two sorts of optimization criteria are met in the literature on automatic scheduling of home appliances: in Conejo et al. (2010), Ferreira et al. (2012), Lujano-Rojas et al. (2012), Volkova et al. (2014) user preferences over home appliance use schedules are represented with a utility function, while in Bradac et al. (2014), Mohsenian-Rad and Leon-Garcia (2010), and in Sou et al. (2011) electricity bill is minimized while user preferences impose constraints on feasible schedules. Most papers assume *a priori* knowledge of the utility function, whereas in the present paper it is learned from consumer's actions.

## 3.2 Preference Learning in Smart Environments

Three popular models of human preferences used in the rational choice theory are the utility function attributed to Von Neumann and Morgenstern (2007), the preference relation, and the choice function. A rational decision maker should always choose the best available alternative, but when preferences are learned from noisy observations, some assumptions are made allowing irrational choices with non-zero probabilities (see Fürnkranz and Hüllermeier (2010)), being it the Luce-Shepard rule and mixed multinomial logit model from McFadden and Train (2000) for utility functions, or the Gaussian process model by Chu and Ghahramani (2005) and its extensions, like those used by Peters (2015), for preference relations. Reinforcement learning (RL) (see Sutton and Barto (1998)) is used to reveal probabilistic choice rules in stochastic environments. An RL-based technique was used by Khalili et al. (2010) to learn user preferences on lightning control in a smart home from her actions, while Peters (2015), Reddy and Veloso (2011) learn customers' electricity tariff selection behavior.

Several supervised learning techniques trained on synthetic data were used by Li et al. (2011) to predict directly stated user comfort level (utility function) on appliance use scenarios, and Manna et al. (2012) use a non-parametric regression to predict thermal comfort learning on the stated preference samples.

In contrast, we do not need a user to fill questionnaires or state her preferences explicitly, but reconstruct preferences from the history of her actions. A similar approach was used by Shann and Seuken (2013) to learn thermal comfort preferences.

## 3.3 Alternative Machine Learning Techniques

We have relatively small observation count (for daily to weekly appliance use) while having a big number of predicting variables (at least equal to the size of the scenario space $A$). Hence, a multiple regression model fitted by ordinary least squares may result in unstable solutions. Linear regression models tend to show low variance having higher bias than more sophisticated models with higher variance and smaller bias, e.g., regression trees. The following popular machine learning tools were chosen to compare their efficiency with that of the proposed Bayesian learning algorithm: $k$ nearest neighbors ($k$NN) by Altman (1992), random forests by Breiman (2001), gradient boosting regression trees by Friedman (2001) (namely, XGBoost algorithm suggested by Chen and He (2015)), support vector regression machines by Smola and Vapnik (1997), and partial least squares (PLS) regression.

# 4. SIMULATION OF CONSUMER BEHAVIOUR

## 4.1 General Model

*A priori* knowledge of the general shape of a comfort level function $d(s, z)$ helps to learn it efficiently. We suggest using detailed time-domain computer simulations of rational consumer behavior to limit the variety of realistic consumer comfort level functions and to parameterize $d(s, z)$ with just a few attributes. Attribute profiles then give rise to hypotheses for the Bayesian inference. The same models generate source data for algorithms of preference learning, as real-world data is unavailable at the moment.

At each moment of time $t$ the state of the system (the smart home environment) is determined by the vector $x(t)$, which includes components of the state of each home appliance, current environment attributes (daytime, day of week, outdoor temperature, etc.), and attributes of consumer state: being at home or not, being hungry or tired, etc.

The system dynamics is driven by the difference equation $x(t + 1) = f(x(t), u(t))$, where $u(t)$ is the vector of control chosen by the consumer at time moment $t$. Control includes all actions with home appliances (changing the conditioner setting, switching lights on and off, etc.), but also going from one room to another or changing the body position, which is important for lighting control in Khalili et al. (2010).

If $d_0(x)$ is the instantaneous consumer comfort level, the consumer's goal function is written as

$$F := \sum_{t=1,\ldots,T} [d_0(x(t) - c_0(x(t), t)],$$

where $c_0(x, t)$ is the cost incurred by all home appliances during a single time period $t$. The function $d_0(x)$ is assumed observable, but the initial system state $x(0)$ is known only by the consumer and is just partially observed by the automatic control system.

The customer's problem of optimal home appliance control reduces to the optimal control problem, which can be solved by the exact dynamic programming algorithm (or RL can find a good approximate solution). Both approaches result in a control schedule $u^*(t)$, the system trajectory $x^*(t)$, and the Bellman's value function $V(u, x)$, which gives the valuation of the control action $u$ under system state $x$. According to Bellman's optimality principle,

$$u^*(t) = \operatorname{argmax}_u V(u, x(t)). \tag{2}$$

If the control system knows $x(0)$, it can reproduce the system dynamics, solve Bellman's equation, calculate the value function $V(u, x)$, and give the best recommendation $u^* = \mathrm{argmax}_u V(u, x)$ for system state $x$. The problem is that the control system only partially observes the system state.

A simple link can be established between Bellman's equation (2) and equation (1). At any given time period $t$ the system state $x(t)$ can be thought as a situation $z = x(t)$ with some observable components $y$ and hidden components $\omega$, the control action $u$ becomes scenario $s = u$, while $V(u, x)$ gives utility function $u(s, z)$, which depends both on the situation $z$ and scenario $s$ chosen (in general, the more complex relation can be more applicable.)

*4.2 Model of Rational Breadmaker Control*

Consider a simple model of rational control of a breadmaker with a single baking program. The bread stock $\sigma(\cdot)$ is spent according to the bread eating schedule $e(t)$ (a private variable) and replenished with breadmaker use during the planning period $t = 0, …, T$. Degree of bread freshness $\phi(\cdot)$ decreases over time. Control action reduces to choosing the desired breadmaker finish period $\Delta$, every model run gives one optimal finish period $\Delta_i^*$. The consumer's goal function is written as:

$$F := \sum_{t = 0,…,T} \{d \cdot \phi(t) \min[\sigma(t), e(t)] - a(\Delta - t)p(t)\},$$

where $d$ is consumer's satisfaction from having bread at a meal (private), $a(\tau)$ – breadmaker power consumption, kW·h / period (see Fig. 2), $p(t)$ – electricity price.

The simulation model and the optimal control problem were implemented in MS Excel and in Matlab Simulink.

*4.3 Using Simulation in Preference Learning*

Simulation output (see Fig. 3) gives rise to the following comfort level and cost function expressions for problem (1):

$d(\Delta, x(0)) = d \sum_{t = 0,…,T} \phi(t) \min[\sigma(t), e(t)],$

$c(\Delta, \cdot) = \sum_{\tau = 0,…,K} a(\tau) p(\Delta - \tau).$

For the fixed $z$ and for $\Delta$ running through a single day the relation $d(\Delta, z)$ is saw-toothed with at most three peaks. The peak heights $d_1, d_2, d_3$, locations $\Delta_1, \Delta_2, \Delta_3$, and the slope $\alpha$ are determined by the system state including hidden factors. We can use this insight to approximate $d(\Delta, z)$ for fixed $z$ with the function of these seven parameters (see Fig. 4).

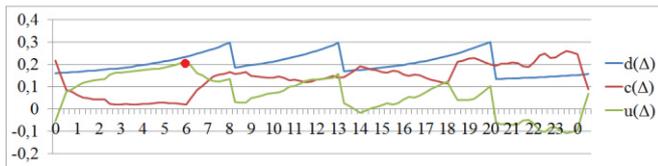

Fig. 3. A typical simulation output (24 hour timeframe): $d(\Delta, \cdot)$ is comfort level under scenario $\Delta$, $c(\Delta, \cdot)$ is the cost of scenario $\Delta$, $u(\Delta, \cdot)$ is the consumer's utility function.

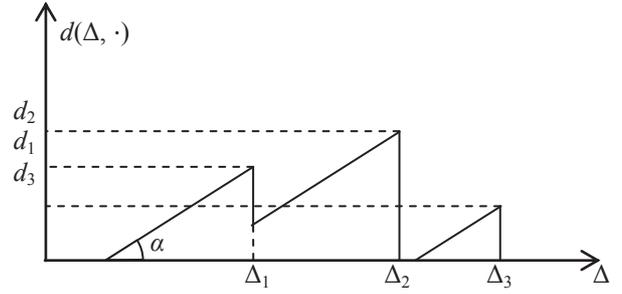

Fig. 4. Approximation of daily comfort level function with seven numeric parameters

## 5. ALGORITHM OF BAYESIAN LEARNING

In the Bayesian framework (see Downey (2013)), $d(\Delta, z)$ as a function of $\Delta$ under fixed $z$ is called a *hypothesis*. Above $d(\Delta, z)$ is described with the vector $\omega = (d_1, d_2, d_3, \Delta_1, \Delta_2, \Delta_3, \alpha)$, so the set of admissible hypotheses $\Omega$ includes all possible combinations of $y$ and $\omega$. Evidence from the training set is used to transform *a priori* probabilities of admissible hypotheses into *a posteriori* probabilities according to the Bayes formula. The algorithm assigns each hypothesis the *likelihood ratio* to reflect the level of its compliance to data. *A posteriori* conditional probabilities $p(\omega|y)$ are hypotheses' likelihood ratios normalized to unity for each observable situation $y$.

Let $r^i$ be the list representing $i$-th observation in the training set, which includes the scenario $\Delta^i$ chosen, observed situation $y^i$ and costs $c^i(\Delta)$ of all alternative scenarios. Let

$$\Delta^*\left(y^i, c^i(\cdot), \omega\right) := \mathrm{argmax}_{\Delta \in A} \left[d\left(\Delta, \left(y^i, \omega\right)\right) - c^i(\Delta)\right]$$

be the optimal scenario chosen under hypothesis $(y^i, \omega)$.

Then, given probability distributions $p(\omega|y)$, situation $y^i$ and costs $c^i(\Delta)$, the best prediction is the *weighted median*

$$\Delta^*\left(y^i, c^i(\cdot)\right) := \min\left\{\Delta : \sum_{\omega: \Delta^*\left(y^i, c^i(\cdot), \omega\right) \leq \Delta} p\left(\omega|y^i\right) \geq 0.5\right\},$$

which minimizes the mean absolute error (MAE) under the assumption that $p(\omega|y)$ is the true hypotheses' distribution.

For every hypothesis $(y, \omega) \in \Omega$ the likelihood ratio is calculated as $L(y, \omega|r^i) = \exp(-\gamma \pi(y, \omega|r^i))$, where $\gamma$ is a tuning parameter (penalty sensitivity), while $\pi(y, \omega | r^i)$ is the function, which penalizes hypothesis $(y, \omega)$ for non-compliance with the real consumer's choice $\Delta^i$:

$$\pi\left(y, \omega|r^i\right) := K\left(y^i, y\right) \cdot$$
$$\cdot \sum_{\Delta \in A} \max\left[0; d\left(\Delta, (y, \omega)\right) - c^i(\Delta) - d\left(\Delta^i, (y, \omega)\right) + c^i\left(\Delta^i\right)\right].$$

The Gaussian kernel function $K(y, y') = \exp(-\beta \rho(y, y')^2)$ boosts learning speed by employing smoothness of $d(\Delta, (y, \omega))$ in $y$: not just the probability of hypotheses for situation $y^i$ are updated but also for its neighbors. Here $\rho(y, y')$ is weighted $L_1$ metric, $\beta$ is sensitivity to distance.

The *a posteriori* likelihood ratio is calculated as follows:

$$L(y,\omega|r^1,\ldots,r^n) = L(y,\omega|r^1) \times \ldots \times L(y,\omega|r^n) \times L_0(y,\omega) =$$
$$= \exp\left(-\gamma \sum_{i=1\ldots n} \pi(y,\omega|r^i) + K_0 \pi_0(y,\omega)\right),$$

where $\pi_0(y,\omega)$ is the penalty, which forms *a priori* probability distribution, and $K_0$ is the weight of priors.

The performance metric of the whole learning process is real MAE calculated as $\sum_{i=1}^{n}\left|\Delta^i - \Delta^*(y^i, c^i(\cdot))\right|$ over the testing observation set.

## 6. COMPUTATIONAL EXPERIMENT SETUP

To generate the data for learning algorithms' training the price schedule $p(t)$ was modeled with a standard three-zonal tariff (expensive evening and cheap night) perturbed by the random walk. The breadmaker control simulation was run with different strength of random walk term: Low, Medium, and High (153, 231, and 127 observations respectively).

The model demonstrates relatively simple rational behavior. The new bread is baked when the stock goes short, and the program finish time is chosen to balance the delight of having the hot bread for the chow time with savings from the low tariff. The typical finish time is 6:00 (a cost-efficient choice exploiting the night tariff), sometimes the bread is ordered to AM $8^{00}$–$9^{00}$ (the breakfast time) and sometimes to PM $1^{00}$–$2^{00}$ (the lunch time). The breadmaker is never run to finish in the evening due to high tariffs.

The predicted variable is program finish time. Every observation $i$ corresponds to one appliance use. Predictors are stock history for today (for 96 periods), baking cost for today and tomorrow (for 182 periods), and the time since the last appliance use.

We build the hypothesis space $\Omega$ for the Bayesian learning algorithm by varying the peak location within the range of two hours ($7^{30}$-$9^{30}$, $12^{30}$-$14^{30}$, and $19^{30}$-$21^{30}$ for three peaks) with the step 30 minutes (5 levels), peak height from ¢9 to ¢14 with step ¢0.3 (20 levels), and slope $\alpha$ from ¢0.5/period to ¢0.9/period with step ¢0.1 (5 levels). Significant observed factors $y$ are the bread stock $\sigma(\Delta_0)$ at the moment of breadmaker run (4 levels from 0 kg to 0.6 kg) and the weekend indicator (0=weekday, 1=weekend). Thus, the table of hypotheses contains $4\cdot 10^7$ elements. Parameters $\beta$ and $\gamma$ are tuned to minimize MAE on the training set ($\beta = 5$, $\gamma = 5$). The algorithm (implemented in Delphi framework) is available at www.dropbox.com/s/5dn8mnptqekilue/Breadmaker.zip together with datasets and settings.

We compare our algorithm with popular regression tools (see Section 3.3). For testing purposes the cross-validation technique is employed with 5-folds and one hold-out sample with 80:20 ratio and 20% last bread-maker runs are selected for the testing set. MAE is used as a main performance metric.

## 7. RESULTS

All learning algorithms were trained on the Medium price volatility dataset. Results are shown in Table 1.

Our Bayesian learning algorithm wins with MAE=0.39 hours (2$^{nd}$ column of Table 1). XGBoost gives the second best MAE = 0.58 hours and alternatives are significantly worse. XGBoost also generalizes well to the other datasets giving the accuracy comparable with the Bayesian learning algorithm (See 1$^{st}$ and 3$^{rd}$ columns of Table 1).

Bayesian learning learns fast (which is an important aspect of customer's adoption of smart appliances, e.g., Nest Learning Thermostat). As shown in Figure 5, for $n = 10$ observations in the training set it gives MAE=0.7 hours on the testing set and MAE<0.6 (the accuracy of the best alternative) for $n \geq 100$.

An open question is robustness of the above algorithm with respect to the model of consumer's behavior. It could be verified by cross-validation with alternative behavioral models (which could be a good exercise for students).

**Table 1. Comparison of algorithms. Learning on the data set with Medium price volatility (231 observations) with 5-fold cross validation. Testing datasets with different price volatilities were selected to verify generalizability.**

| Price volatility for the testing set | Low | Medium | High |
|---|---|---|---|
| Algorithm | Mean absolute error, hours | | |
| kNN | 0.80 | 0.95 | **0.85** |
| SVM regression | 0.65 | 0.88 | 0.92 |
| XGBoost | **0.34** | 0.58 | 0.86 |
| PLS regression | 0.87 | 1.13 | 0.99 |
| OLS linear regression | 2.93 | 2.93 | 5.72 |
| Random forest | 0.59 | 0.89 | 0.96 |
| Mean | 1.01 | 1.16 | 1.14 |
| Ridge regression | 0.94 | 0.93 | 0.89 |
| Lasso regression | 0.97 | 1.55 | 1.67 |
| **Bayesian learning** | **0.32** | **0.39** | **0.83** |

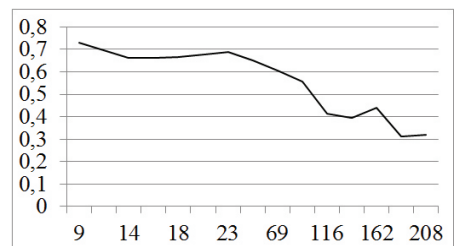

Figure 5. MAE (hours) of Bayesian learning algorithm vs size of the training set. $n$ observations are randomly selected for the training set from the total of 231 observations ("Medium" data set), all other observations are left for the testing set.

## 6. CONCLUSIONS

The distinctive feature of the proposed approach to home appliance control automation for demand-side management is the focus on learning consumer's response to electricity price signals. The recommender / automatic control system suggests scenarios of home appliance use according to the probability distributions of possible comfort level functions learned by the Bayesian inference algorithm.

Efficiency of the approach is verified on an example of a small-scale household appliance (a breadmaker), but the approach can also be applied to improve air heating/cooling by introducing the price of indoor temperature shift beyond the range of consumer's comfort. Thus, consumer's comfort sensitivity to price changes becomes the subject of learning.

Author 1 acknowledges Russian Science Foundation grant 16-19-10609 (covers simulation and learning models).